# A Comparative Study on TF-IDF feature Weighting Method and its Analysis using Unstructured Dataset

Mamata Das, Selvakumar Kamalanathan and Pja Alphonse

*NIT Trichy, 620015, Tamil Nadu, India*

**Abstract**

Text Classification is the process of categorizing text into the relevant categories and its algorithms are at the core of many Natural Language Processing (NLP). Term Frequency-Inverse Document Frequency (TF-IDF) and NLP are the most highly used information retrieval methods in text classification. We have investigated and analyzed the feature weighting method for text classification on unstructured data. The proposed model considered two features N-Grams and TF-IDF on the IMDB movie reviews and Amazon Alexa reviews dataset for sentiment analysis. Then we have used the state-of-the-art classifier to validate the method i.e., Support Vector Machine (SVM), Logistic Regression, Multinomial Naïve Bayes (Multinomial NB), Random Forest, Decision Tree, and k-nearest neighbors (KNN). From those two feature extractions, a significant increase in feature extraction with TF-IDF features rather than based on N-Gram. TF-IDF got the maximum accuracy (93.81%), precision (94.20%), recall (93.81%), and F1-score (91.99%) value in Random Forest classifier.

**Keywords 1**
TF-IDF, N-Gram, Text classification, Feature weighting, Information retrieval.

## 1. Introduction

Nowadays, we are dealing with a huge volume of data. There has been a considerable rise in social platform giants such as Twitter, WhatsApp, Facebook, YouTube, and Instagram [1] and IMDB movie review platforms thus proving them to be a massive amount of big data. We must use various techniques of data mining for finding potentially useful patterns from this huge data before we can extract necessary information from that. Sentiment analysis is the practice to classify various samples of related text video and audio [2] into overall positive, negative, and neutral categories [3] using different algorithms.

Mankind is facing a terrible experience due to this covid-19 pandemic. The whole world is losing the balance from each and every sector, directly or indirectly related to financial issues. We are sharing our opinion on social platforms like Facebook. Various types of tweets are being tweeted on platforms like Twitter. Big data techniques may be useful to classify the huge amount of various types of data like textual data, audio data, visual data, and image data.

Online shopping websites have a great significance on our daily lives as human beings in the twenty-first century are very busy and productive. We share our reviews about our satisfaction after buying products from numerous shopping giants like eBay, Amazon, Flipkart [4]. Those reviews have two-fold effects on the shopping ecosystem: i) It is a basic human tendency before we buy the product for our own needs, we check those reviews posted by other buyers, (ii) manufacturers also put a high-level focus on those reviews to make their business smoother [5]. Most people are shopping online in this Covid-19 pandemic situation and they depend a lot on those textual tweets or reviews before buying for themselves [6]. This may be considered an integral part of the habits of human beings. Those reviews

---

generally contain human sentiment attached with the reviews of the product or food or movie or political views etc. Sentiment analysis (SA) may be considered as a ubiquitous model to analyze that huge volume of tweets or textual data. Short comments or reviews about a newly launched movie play a vital role in the success or failure of the movie[7]. People generally investigate from internet movie databases like RottenTomatoes, Yahoo movies to get informed about movie plot, storyline, music, cast, production, marketing, release date, and sequels, etc. Textual reviews of critics along with comments of viewers are very essential parts of a movie to make it successful in the movie market.

Some of the most popular feature extraction techniques are NLP based features, Bag-of-Words, TF-IDF, etc. Feature weighting is one of the many techniques that we use to find out which features are important, and which are not from a document. Calculating term frequency and inverse document frequency is an important feature weighting technique for feature extraction. In this paper, we analyzed and compare the two feature extraction techniques N-Gram and TF-IDF on two datasets (IMDB movie reviews and Amazon Alexa reviews dataset) of sentiment analysis.

We have applied various classification methods to compare and pick out which among the six feature extraction techniques is most suitable. We have gone through some steps for data preprocessing. After this preprocessing step, we have extracted two features from the text. We have addressed the effects of various features like TF-IDF as well as n-grams for n = 2 on the performance of SA.

Distinct feature weighting algorithms based on grams and TFIDF numeric statistics motivated us to pursue this research work. Although the current algorithm on which we want to study is based upon information retrieval and text mining, we want to verify the algorithm on some recent datasets which deal with textual data.

Now we mention the structure of the paper. An overview of the literature survey has been presented in section 2, the methodology has been shown in section 3. The performance of the classifiers and experimental results are shown in section 4. Section 5 presents discussions and conclusions and, Section 6 contains future work.

## 2. Related Works

Social and business applications have triggered a wide range of increasing rate of research on sentiment analysis over the last 15 years [8,9]. People share their experience in the form of textual notes which in turn helps us to know about the advantage and disadvantages of the product or food or movie or political views [10]. Various e-commerce companies like Amazon, Flipkart are improving the services as well as products on the basis of various texts, tweets, sentiments, blogs, reviews.

Deep learning-based tools, such as Word2vec, helps us to represent the feature words. Authors have shown this kind of method for text classification analysis. The weight of the feature words has been calculated by using the improved version of the TF-IDF method [11]. They have achieved higher accuracy in terms of classification compared with the classical TFIDF model as indicated by their results.

Various music genres like Pop, hip-hop, county, metal, and rock have been used to study the sentiment analysis by numerous researchers as indicated in [12]. They have classified those music genres using various classification models and devised a model to predict genres with lyrics based on various parameters such as various characteristics of lyrics e.g. its length, frequent words in lyrics, length of words.

Feature extraction methods are an essential step while predicting the subcellular location of proteins in the field of bioinformatics, genome annotation, and computational biology [25].

Keyword ranking on 50 recent image captions based on Instagram users has been successfully shown in [13]. Here TF-IDF calculated the ranking of keywords as mentioned by them. The main topic of a user is defined by the highest ranking of keywords. As shown in their experiments, TF-IDF finds and ranks the keywords of users' image captions from the Instagram dataset.

The application of TF-IDF has beautifully been presented in terms of comments mining by the authors as mentioned in [14]. They have studied the TF-IDF model and proposed adjustments for bias. They have presented their work on highly correlated content which results in dependency. They have verified their proposal with seven Facebook fan page data covering different domains including news, finance, politics, sports, shopping, and entertainment.

Unigram, bigram, trigram have been used to classify the sentiment on the IMDB dataset using various deep learning methods [7]

The paper concerns with the automatic classification of bengla documents. They have proposed a categorization system, a support vector machine for the classification of documents, and TFIDF is used for feature selection [15].

Authors have analyzed and research feature word weight which is used in unstructured data classification of big data [16]. They modified the traditional TFIDF algorithm formula, adding the concept of intra-class dispersion, excluding the inner impact to disturb characteristic.

An n-gram based TF-IDF approach has been mentioned in [24] to detect hate speech and offensive language on social media like Twitter. They have considered three labels on the Twitter dataset to indicate classes: hateful(0), offensive(1), and clean(2). Tweet classification has been done using the Term frequency and document frequency.

[17, 18] They have proposed some research on sentiment analysis on Twitter data set and also discussed the effect of the pre-processing techniques on it.

Other methods have also been shown in [8,19,20] to detect Sentiment from textual data. Researchers have proposed various modifications to the classical TF-IDF models and proved their efficiency on different types of datasets. Semantic information may be useful for further improvement of the performance of the TF-IDF model.

## 3. Methods
### 3.1. Dataset

We have taken the well-known publicly available movie review dataset to evaluate the performance of the proposed methods. We have performed our experiment on the IMDB movie review dataset proposed by [21], which consists of 50,000 reviews, which contain 25,000 reviews labeled as positive and 25,000 reviews labeled as negative. There are benchmark datasets available for analysis of sentiment on movie reviews. IMDB is one of them. We also conducted experiments using the Amazon Alexa user's reviews dataset. The Amazon Alexa users reviews dataset contains 3000 reviews labeled with positive and negative.

### 3.2. Data Pre-processing Techniques

Data preprocessing is an important issue while processing raw datasets. In this step, we transformed, or Encoded the data so the machine can easily parse it. We have gone through some steps for data preprocessing.

### 3.2.1. Missing Values

Missing values are a common incident in the dataset, and we have a different scheme for handling them. Here, we have eliminated rows with missing data.

### 3.2.2. Tokenization

Tokenization is a step where chunks of text can be splits into sentences and sentences can be further split into words. To give an idea, Let's take an example : "He who does not research has nothing to teach" and after tokenization: {'He', 'who', 'does', 'not', 'research', 'has', 'nothing', 'to', 'teach'}

### 3.2.3. Normalization

In data preprocessing, normalization is a highly overlooked step where the text is transformed into a canonical form. There are series of tasks performed to achieve normalization like transforming all text to either lower or upper case, removing punctuation, converting numbers to their equivalent words, and so on. It keeps all words on an equal scale to move the processing forward evenly. Text normalization is important for noisy texts such as text messages, product reviews, social media comments, and comments to blog posts where misspellings, abbreviations, and use of oov (out-of-vocabulary) words are usual. For example, before and after normalization: b4 to before, :) to smile and 2moro to tomorrow.

### 3.2.4. Stemming

Stemming is the process of changing a non-root word to its root form. The stemming is done by eliminating affixes (infixes, suffixed, prefixes, circumfixes) from the words. For example, studies, studied, studying, studied to study and connect, connected, connections, connections, connects to connect.

### 3.2.5. Lemmatization

Lemmatization is an important aspect of natural language processing which is used to remove inflections and map a word to its base or dictionary form. Here is an example of lemmatization in action: goose, geese to a goose.

### 3.2.6. Remove Stop Words

Stop words are commonly used words in the English language which are filtered out before or after processing natural language data. Some of the stop words are "the", "a", "are", "at", "in", "and", etc. So, these need to be allayed for better prediction.

### 3.2.7. Noise removal

Real-world data is generally noisy, incomplete, and inconsistent due to fault in data collection, data transmission errors, or error during data entering. We remove a specific row that has a null value for a particular column or a feature. The noise removal is done very carefully, failing to do this, we might get an inaccurate and faulty conclusion or leads to decreased accuracy.

### 3.3. Feature Extraction

### 3.3.1 N-Gram

In supervised machine learning algorithms, N-Grams are used as a method of feature extraction selection. N grams are a set of n-consecutive tokens from a given tweet or speech or text or blog. terminology exists for n =1,2,3 as unigram, bigram, trigram, respectively. Let's think over the sentence "Research is creating new knowledge". Let us consider N=2 then it will bring us "Research is", "is creating", "creating new", "new knowledge". In our proposed methodology, we have used n = 2 i.e., bigram.

### 3.3.2 TF-IDF

TF-IDF is a statistical model to evaluates the significance of words in a document pool. It may be calculated by multiplying two metrics: TF-matrix which is a two-dimensional matrix and IDF which is a one-dimensional matrix. Various refinements have been proposed by several researchers to improve the classical TF-IDF model. Suppose we are given a corpus $\{D_1, D_2, …, D_N\}$ with N documents. Let $D_i$ denote any arbitrary document for $1 \leq i \leq p$. Let $D_j = \{t_1, t_2, …, t_q\}$ be a document containing q terms. Let ti be an arbitrary feature word. If n(i,j) denotes the number of appearances of ti in document Dj then the term frequency of ti concerning Dj, denoted by TF(i,j), may be defined by the equation:

$TF(I, j) = \frac{n(I,j)}{n(1,j)+ n(2,j)+\cdots+ n(q,j)}$. Let $n_i$ denotes the total number of documents where ti appears at least once. Then inverse document frequency of $t_i$ concerning the corpus, denoted by $IDF_i$ may be defined as $IDF_i = \log(\frac{N}{ni})$. IDF(i) can be stated as the ratio of the total number of documents in hand with the number of documents where feature words exist. It indicates the distribution of $t_i$ in the whole corpus. TF-IDF is a vector space model (VSM) that determines how much a particular word is relevant with respect to a document. Let us have a document that contains hundreds of words. Let the frequency of the particular word "research" is 5. Then its TF will be 5/100 = 0.05. Now suppose, for example, there are 40000 relevant documents. Among them, the words "research" shows up 400 times. Then Inverse-document-frequency is calculated as 40000/400=100. Thus we can find the value of TF-IDF as = 5.

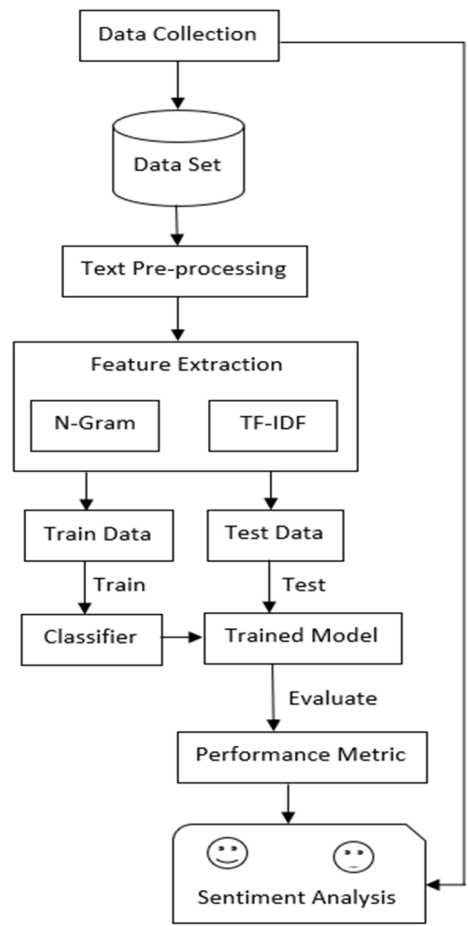

**Figure 1**: The Framework of Proposed Methodology

## 3.4. Performance Parameters

In our research, the key realization is that not all correct or incorrect matches hold equal value. A single metric will not tell the whole evaluation of classification. Therefore, we have used accuracy, recall, precision and, F1 score as performance metrics that are discussed in the next section.

### 3.4.1. Accuracy (A)

Accuracy may be defined as the ratio of cardinality of correct prediction with cardinality of actual prediction. If we have high accuracy or near too high, then we can say our model is best. So, accuracy may be defined as follows:

$$A = \frac{\alpha}{\beta} \qquad (1)$$

Where, α = Cardinality of the prediction which are correct.
  β = Cardinality of actual prediction.

### 3.4.2. Precision (P)

In simple words, precision may be thought of as a measure of classifiers exactness i.e., the proportion of identification that was correct. It is the rate of correctly predicted positive instances to the total predicted positive instances. A low precision value can indicate many False Positives. So, the precision can be defined as follows:

$$P = \frac{\text{Relevant Document} \cap \text{Retrieve Document}}{\text{Retrieve Document}} \qquad (2)$$

### 3.4.3. Recall (R)

Recall may be thought of as a measure of a classifier's completeness which gives us the proportion of actual positive which is identified correctly by the predicted classifier. It is the ratio of correct positive instances and combines with correct positive and wrong positive. It is also called Sensitivity. If recall returns us a low recall value, then it indicates many False Negatives. So, the recall can be defined as follows:

$$R = \frac{\text{Relevant Instances}}{\text{Total Amount of Instances}} \qquad (3)$$

### 3.4.4. F1 Score (F1)

In an uneven class distribution, F1 is a good choice as a performance metric. It is the weighted average of both properties, Recall and Precision. This performance metric takes both false negatives and false positives into the calculation. Differently, we may say that the F1 score carries the balance between recall and precision. So, the F1 score can be defined as follows:

$$\text{F1 score} = 2 * \frac{P * R}{P + R} \qquad (4)$$

Good information retrieval or text classification classifier is expected to be able to provide precision, recall and F1 score values high or near to high.

### 3.5. Execution environments

We have implemented our experimental execution on a Lenovo ThinkPad E14 Ultrabook with Windows 10 Professional 64-bit operating system and 10th Generation Intel Core i7-10510U Processor. The clock speed of the processor is 1.8 GHz with 16G bytes DDR4 memory size.

## 4. Results

**Table 1**
Performance classification using the n-gram method.

| Classifier | IMDB Dataset | | | | Amazon Alexa Reviews | | | |
|---|---|---|---|---|---|---|---|---|
| | A | P | R | F1 | A | P | R | F1 |
| Multinomial NB | 88.14 | 88.20 | 88.14 | 88.14 | 93.02 | 93.51 | 93.02 | 90.49 |
| SVM | 88.54 | 88.57 | 88.54 | 88.54 | 92.54 | 93.10 | 92.54 | 89.50 |
| KNeighbors | 56.52 | 57.08 | 56.52 | 55.74 | 91.75 | 84.45 | 91.75 | 87.95 |
| LogisticRegression | **90.47** | **90.48** | **90.47** | **90.47** | **93.65** | 93.57 | **93.65** | **91.87** |
| Decision Tree | 65.78 | 69.53 | 65.78 | 64.00 | 92.38 | 90.52 | 92.38 | 90.05 |
| Random Forest | 86.37 | 86.41 | 86.37 | 86.37 | 93.49 | **93.92** | 93.49 | 91.42 |

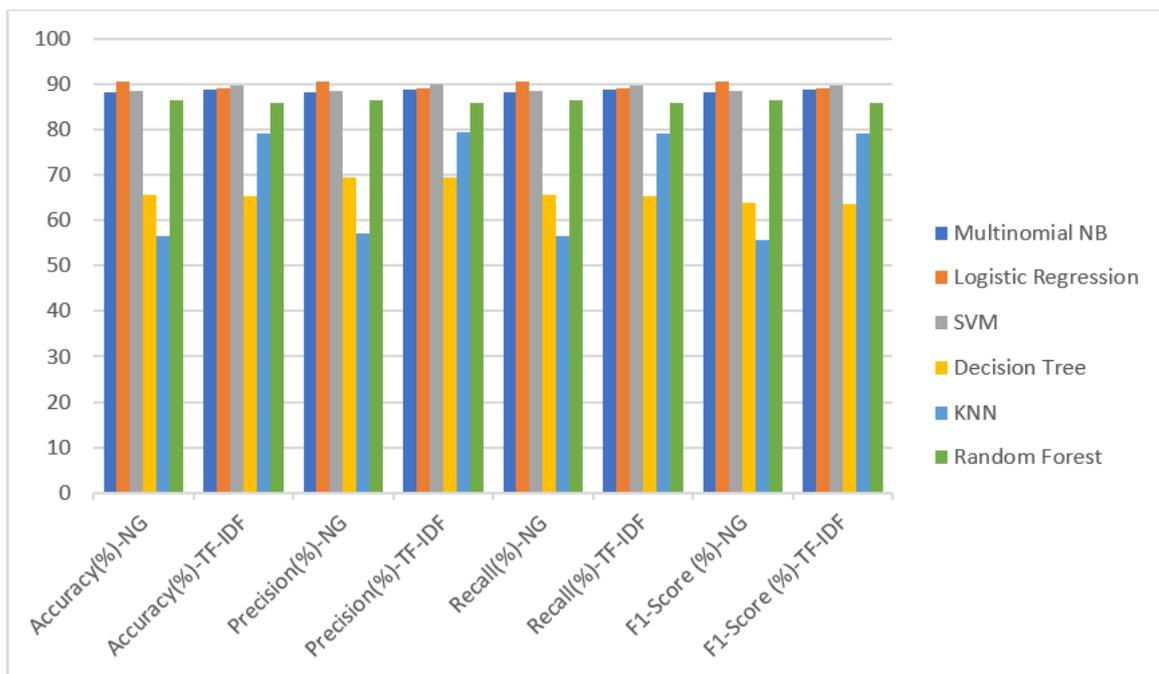

**Figure 2**: Comparison of two approaches (TF-IDF and N-gram) on Movie Reviews Dataset

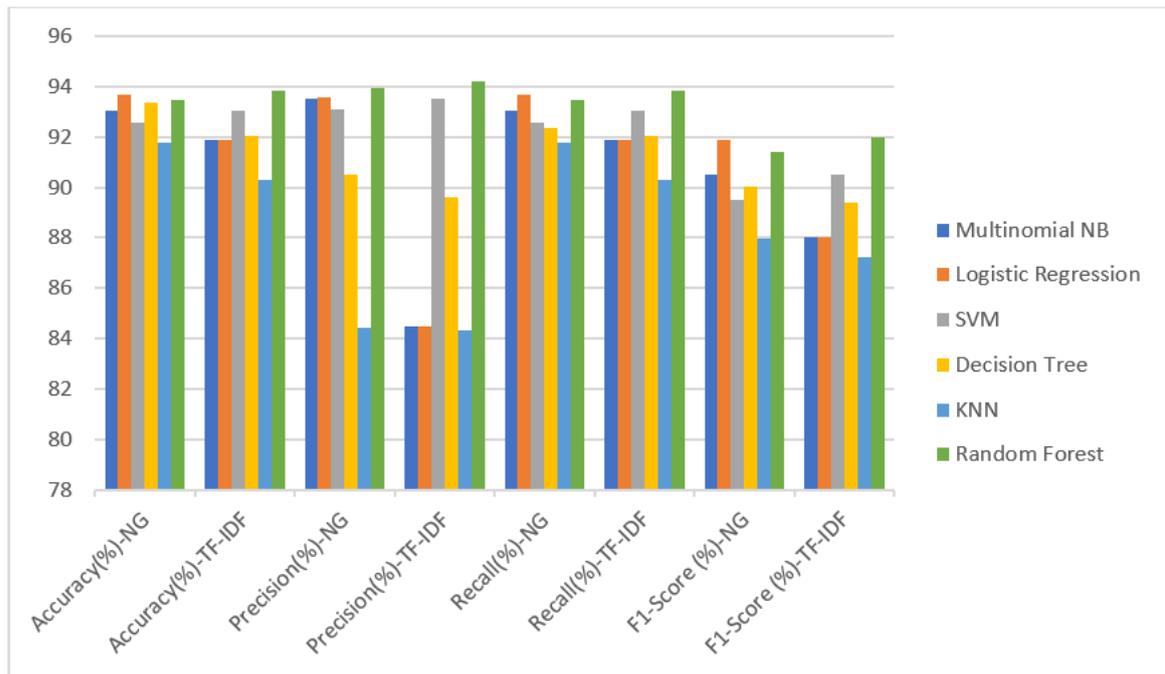

**Figure 3**: Comparison of two approaches (TF-IDF and N-gram) on Amazon Alexa Reviews

**Table 2**
Performance classification using the TF-IDF method.

|  | IMDB Dataset | | | | Amazon Alexa Reviews | | | |
| --- | --- | --- | --- | --- | --- | --- | --- | --- |
| Classifier | A | P | R | F1 | A | P | R | F1 |
| Multinomial NB | 88.73 | 88.77 | 88.73 | 88.73 | 91.90 | 84.46 | 91.90 | 88.03 |
| SVM | **89.87** | **89.90** | **89.87** | **89.87** | 93.02 | 93.51 | 93.02 | 90.49 |
| KNeighbors | 79.15 | 79.50 | 79.15 | 79.08 | 52.50 | 52.39 | 90.32 | 87.23 |
| LogisticRegression | 89.20 | 89.22 | 89.20 | 89.20 | 91.90 | 84.46 | 91.90 | 88.03 |
| Decision Tree | 65.40 | 69.35 | 65.40 | 63.48 | 92.06 | 89.59 | 92.06 | 89.42 |
| Random Forest | 85.80 | 85.80 | 85.80 | 85.80 | **93.81** | **94.20** | **93.81** | **91.99** |

## 5. Discussion and Conclusions

In this paper, we considered two features, such as Bigram and TF-IDF, and analyzed their performance on IMDB movie review and Amazon Alexa reviews dataset for SA. In Table 1, We have taken four Parameters affecting the performance, such as accuracy(A), precision(P), F1 score(F1), and recall(R), and six classification algorithms such as KNN, Logistic Regression, Decision Tree, Multinomial NB, Random Forest, and, SVM to implement N-gram features. In Table 2, We have taken four Parameters affecting the performance, such as accuracy(A), precision(P), F1 score(F1), and recall(R), and six classification algorithms such as Multinomial NB, SVM, Logistic Regression, Random Forest, Decision Tree, and KNN to implement TF-IDF features. Logistic regression outperforms the other classifiers in Ngram features. In the case of TF-IDF, SVM and Random Forest are performing better among all the classifiers. As it can be seen from both the tables, TF-IDF got the maximum accuracy (93.81%), precision (94.20%), recall (93.81%), and F1-score (91.99%) value in the Random Forest classifier, and Fig2 and Fi3 show the comparative review of the performance of the features. Future research direction.

In the future, clustering techniques may be used to improve our current research results. One such clustering technique is a k-means algorithm. We have planned for another experiment of TF-IDF with the temporal k-means algorithm mentioned in [22] which may improve the classification performance.

An alternative for IDF is Inverse gravity moment (IGM) that may be used for term weighting in text classification as mentioned in [23]. Researchers have paid a lot of effort to discover the mathematical model of IGM. Our next motivation may be the study the mathematical model of IGM and compare the performance of improved TF-IDF with TGM.

## 6. References

[1] A. Singh, M N. Halgamuge and B. Moses. "An analysis of demographic and behavior trends using social media: Facebook, Twitter, and Instagram." Social Network Analytics (2019): 87-108.
[2] S. Poria, E. Cambria, D. Hazarika, N. Majumder, A. Zadeh and L-P. Morency. "Context-dependent sentiment analysis in user-generated videos, in: Proceedings of the 55th annual meeting of the association for computational linguistics (volume 1: Long papers)", Association for Computational Linguistics, Vancouver, Canada, 2017, pp. 873—883. doi: 10.18653/v1/P17-1081.
[3] I. Chaturvedi, E. Cambria, R E. Welsch and F. Herrera. "Distinguishing between facts and opinions for sentiment analysis: Survey and challenges", Information Fusion, 44(2018): 65-77
[4] Y. Liu, J W. Bi and Z P. Fan. "Ranking products through online reviews: A method based on sentiment analysis technique and intuitionistic fuzzy set theory", Information Fusion, 36(2017): 149-161
[5] D. Law, R. Gruss and A S. Abrahams. "Automated defect discovery for dishwasher appliances from online consumer reviews", Expert Systems with Applications, 67(2017): 84-94
[6] K. Zhao, A C. Stylianou and Y. Zheng. "Sources and impacts of social influence from online anonymous user reviews", Information & Management, 55(2018): 16-30
[7] Abinash Tripathy, Ankit Agrawal and Santanu Kumar Rath. "Classification of sentiment reviews using n-gram machine learning approach", Expert Systems with Applications, 57(2017): 117-126
[8] B. Liu. "Sentiment analysis and opinion mining", Synthesis lectures on human language technologies, 5.1(2012): 1-167
[9] Erik Cambria, Catherine Havasi and Amir Hussain, "Senticnet 2: A semantic and affective resource for opinion mining and sentiment analysis", In: Proceedings of the 25th International Florida Artificial Intelligence Research Society Conference, FLAIRS-25, 2012, pp.202-207.
[10] B. Agarwal and N. Mittal. "Prominent feature extraction for review analysis: an empirical study", Journal of Experimental & Theoretical Artificial Intelligence, 28(2016):485-498
[11] Cai-zhi Liu, Yan-xiu Sheng, Zhi-qiang Wei and Yong-Quan Yang, Research of text classification based on improved TF-IDF algorithm", 2018 IEEE International Conference of Intelligent Robotic and Control Engineering (IRCE), 2018, pp. 218-222. doi: 10.1109/IRCE.2018.8492945
[12] V. Bangwal, S. Kamalnathan, Y. Mishra and V. Kumawat. "Evolution of Different Music Genres", International Journal of Engineering and Advanced Technology (IJEAT), 9(2019):5138-5143
[13] Bernardus Ari Kuncoro and Bambang Heru Iswanto, "TF-IDF method in ranking keywords of Instagram users' image captions," 2015 International Conference on Information Technology Systems and Innovation (ICITSI), Bandung, 2015, pp. 1-5. doi: 10.1109/ICITSI.2015.7437705
[14] Inbal Yahav and O. Shehory and D. Schwartz. "Mining With TF-IDF: The Inherent Bias and Its Removal", EEE Transactions on Knowledge and Data Engineering, 31(2019):437-450
[15] Md Saiful Islam, Fazla Elahi Md Jubayer and Syed Ikhtiar Ahmed. "A support vector machine mixed with TF-IDF algorithm to categorize Bengali document", 2017 international conference on electrical, computer and communication engineering (ECCE), 2017, pp.191-196.
[16] Aizhang Guo and Tao Yang. "Research and improvement of feature words weight based on TFIDF algorithm", 2016 IEEE Information Technology, Networking, Electronic and Automation Control Conference, Chongqing, 2016, pp. 415-419, doi: 10.1109/ITNEC.2016.7560393.

[17] Z. Jianqiang and G. Xiaolin. "Comparison Research on Text Pre-processing Methods on Twitter Sentiment Analysis," in IEEE Access, 5(2017): pp. 2870-2879. doi: 10.1109/ACCESS.2017.2672677.
[18] T. Singh and M. Kumari. "Role of Text Pre-processing in Twitter Sentiment Analysis", Procedia Computer Science,89(2016):549-554
[19] A. Dey, M. Jenamani, J J. Thakkar. "Senti-N-Gram: An n-gram lexicon for sentiment analysis", Expert Systems with Applications, 103(2018):92-105.
[20] F. Aisopos, G. Papadakis, and T. Varvarigou. "Sentiment analysis of social media content using N-Gram graphs", In: Proceedings of the 3rd ACM SIGMM international workshop on Social media (WSM '11). Association for Computing Machinery, New York, NY, USA, (2011) 9–14. doi: https://doi.org/10.1145/2072609.2072614
[21] A L. Mass, R E. Maas, Daly, and T. P Peter and D. Huang and Y. Ng and P. Christopher, "Learning Word Vectors for Sentiment Analysis", Proceedings of the 49th Annual Meeting of the Association for Computational Linguistics: Human Language Technologies, Association for Computational Linguistics, Portland, Oregon, USA, 2011, pp. 142--150.
[22] S. Kamalanathan , L. Sai Ramesh and A. Kannan . "Enhanced K-Means Clustering Algorithm for Evolving User Groups", Indian Journal of Science and Technology, 8(2015):1-8
[23] K. Chen and Z. Zhang and J. Long and H. Zhang. "Turning from TF-IDF to TF-IGM for Term Weighting in Text Classification", Expert Systems with Applications, 66(2016)245-260.
[24] A. Gaydhani, V. Doma, S. Kendre, L. Bhagwat. "Detecting Hate Speech and Offensive Language on Twitter using Machine Learning: An N-gram and TFIDF based Approach", arXiv preprint arXiv:1809.08651 (2018)
[25] L. Du, Q. Meng , Y. Chen and P. Wu. "Subcellular location prediction of apoptosis proteins using two novel feature extraction methods based on evolutionary information and LDA", BMC bioinformatics, 21(2020)1--19.